\documentclass[10pt,twocolumn]{article}

\usepackage[T1]{fontenc}
\usepackage[utf8]{inputenc}
\usepackage[english]{babel}

\usepackage[
  top=2cm, bottom=2cm,
  left=1.5cm, right=1.5cm,
  columnsep=0.6cm
]{geometry}

\usepackage{lmodern}
\usepackage{microtype}

\usepackage{amsmath}
\usepackage{amssymb}

\usepackage{graphicx}
\usepackage{float}
\usepackage[labelfont=bf,font=small]{caption}
\usepackage{subcaption}

\usepackage{booktabs}
\usepackage{multirow}
\usepackage{array}

\usepackage{listings}
\usepackage{xcolor}
\definecolor{codebg}{HTML}{F5F5F5}
\definecolor{codeframe}{HTML}{CCCCCC}
\lstdefinestyle{prompt}{
  basicstyle=\ttfamily\footnotesize,
  backgroundcolor=\color{codebg},
  frame=single,
  rulecolor=\color{codeframe},
  framerule=0.5pt,
  framesep=6pt,
  breaklines=true,
  columns=fullflexible,
  showstringspaces=false,
  upquote=true,
  xleftmargin=0pt, xrightmargin=0pt
}

\usepackage{pifont}
\newcommand{\cmark}{\ding{51}}
\newcommand{\xmark}{\ding{55}}
\usepackage{hyperref}
\hypersetup{
  colorlinks=true,
  linkcolor=blue!70!black,
  citecolor=blue!70!black,
  urlcolor=blue!70!black
}

\usepackage[
  backend=biber,
  style=numeric,
  sorting=none,
  maxbibnames=3,
  minbibnames=1
]{biblatex}
\title{%
  \Large\bfseries
  Depression Risk Assessment in Social Media\\[2pt]
  via Large Language Models
}

\author{%
  Giorgia Gulino\textsuperscript{1}\quad
  Manuel Petrucci\textsuperscript{1}\\[4pt]
  \small\textsuperscript{1}Guglielmo Marconi University,
  Department of Human Sciences, Rome, Italy\\[2pt]
   \small Corresponding: \texttt{g.gulino1@studenti.unimarconi.it}
}

\date{}

\begin{document}

\maketitle
\thispagestyle{empty}

\begin{abstract}
Depression is one of the most prevalent and debilitating mental health conditions worldwide, frequently underdiagnosed and undertreated. The proliferation of social media platforms provides a rich source of naturalistic linguistic signals for the automated monitoring of psychological well-being. In this work, we propose a system based on \textit{Large Language Models} (LLMs) for depression risk assessment in Reddit posts, through \textit{multi-label} classification of eight depression-associated emotions and the computation of a weighted severity index. The method is evaluated in a \textit{zero-shot} setting on the annotated \emph{DepressionEmo} dataset ($\approx6{,}000$ posts) and applied \textit{in-the-wild} to $469{,}692$ comments collected from four subreddits over the period 2024--2025. Our best model, \texttt{gemma3:27b}, achieves micro-F1 = 0.75 and macro-F1 = 0.70, results competitive with purpose-built fine-tuned models (BART: micro-F1 = 0.80, macro-F1 = 0.76). The in-the-wild analysis reveals consistent and temporally stable risk profiles across communities, with marked differences between \textit{r/depression} and \textit{r/anxiety}. Our findings demonstrate the feasibility of a cost-effective, scalable approach for large-scale psychological monitoring.
\end{abstract}


\section{Introduction}
\label{sec:intro}

Depression is one of the most prevalent and debilitating mental disorders globally, recognized by the WHO as a leading cause of disability~\cite{GBD2018,who2017}. Despite the availability of effective treatments, a substantial proportion of cases remains undiagnosed or insufficiently treated, particularly among young adults~\cite{GBD2018}. Early detection of depressive distress signals is therefore of critical importance for timely intervention.

The pervasive digitalization of daily life has made social media a privileged observatory for studying psychological behaviors and emotional expressions. Individuals routinely share thoughts and emotional states online, constituting an indirect yet relevant source of information on psychological well-being~\cite{liu2022,Conway2016,Naslund2019}. Computational psychology research has demonstrated that textual content published on platforms such as Reddit and Twitter contains linguistic markers of emotional distress, enabling the early identification of depressive risk~\cite{DeChoudhury2013,Eichstaedt2018,AlMosaiwi2018}.

In parallel, advances in Natural Language Processing (NLP) have made it possible to systematically analyze large volumes of online text. The introduction of Transformer-based architectures~\cite{devlin2019} and, more recently, \textit{Large Language Models} (LLMs) has opened new possibilities: these models learn deep linguistic representations from massive text corpora, developing a sensitivity to emotional nuance that is difficult to capture with classical approaches~\cite{gadzama2024,Jin2025}.

Compared to fine-tuned models, LLMs offer a significant operational advantage: they can be deployed in \textit{zero-shot} mode without requiring domain-specific labeled data, drastically reducing development costs while improving adaptability to rapidly evolving linguistic contexts~\cite{ohse2024zero}.

This paper makes the following contributions:
\begin{enumerate}
  \item A {weighted depressive severity index} based on eight clinically relevant emotion categories, inspired by the standardized PHQ-9 and BDI-II scales.
  \item A {prompt engineering methodology} for zero-shot multi-label classification of depressive emotions via LLMs.
  \item A {systematic evaluation} of nine locally-run LLMs (0.6B--27B parameters) against fine-tuned baselines from the literature on the \emph{DepressionEmo} dataset~\cite{rahman2024depressionemo}.
  \item An {\textit{in-the-wild} analysis} of $469{,}692$ Reddit posts (2024--2025), examining emotion distributions, risk profiles, and longitudinal trends.
\end{enumerate}

The proposed system is not intended as a replacement for clinical assessment, but as a scalable support and triage tool for large-scale psychological monitoring~\cite{Chancellor2020}, on limited affordable hardware running models in local.

The paper is structured as follows: Section~\ref{sec:related} shows the related works in clinical measurement of depression severity, machine learning applied to depression detection and emotions relevant for depression identification. Section~\ref{sec:method} presents the our approach with the score and the prompt used to guide LLMs in the detection. Section~\ref{sec:experiments} presents two section of experiments: controlled and in-the-wild, which are then discussed in Section~\ref{sec:results}. Finally, Section~\ref{sec:conclusions} poses the conclusion of this work.

\section{Related Work}
\label{sec:related}

This section presents three key areas of prior research that contextualize our approach. First, we review the clinical measurement of depression severity, focusing on standard psychometric scales and the relative clinical weight of specific symptoms (Section~\ref{sec:related_clinical}). Second, we trace the evolution of NLP and machine learning techniques for depression detection, from traditional algorithms to large language models (Section~\ref{sec:related_nlp}). Finally, we examine the literature concerning the specific emotional states associated with depression and their linguistic taxonomies (Section~\ref{sec:related_emotions}).
\subsection{Clinical Measurement of Depression Severity}
\label{sec:related_clinical}

Depression severity is traditionally assessed through standardized psychometric scales. The DSM-5 defines nine core symptoms of major depressive disorder, including depressed mood, anhedonia, sleep and appetite disturbances, fatigue, difficulty concentrating, feelings of worthlessness or guilt, and recurrent thoughts of death~\cite{apa2022,milintsevich2023towards}. Instruments such as the \textit{Patient Health Questionnaire-9} (PHQ-9) and the \textit{Beck Depression Inventory-II} (BDI-II) quantify these symptoms through additive scoring, with established thresholds distinguishing mild, moderate, and severe depression~\cite{oh2024development}.

Although each symptom nominally contributes equally to the total score, clinical practice assigns de facto greater weight to specific indicators. In particular, persistent hopelessness is recognized as a strong risk factor for suicidal ideation and attempts~\cite{ribeiro2018depression}, while suicide intent warrants immediate clinical intervention. These clinical considerations informed the weighting scheme adopted in the present work.

\subsection{NLP and Machine Learning for Depression Detection}
\label{sec:related_nlp}

Early computational approaches to depression detection in social media relied on traditional machine learning methods (SVM, logistic regression) operating on hand-crafted linguistic features: frequency of emotionally negative words, first-person pronouns, and absolutist expressions~\cite{DeChoudhury2013,AlMosaiwi2018,Guntuku2017}. While effective in controlled settings, these methods exhibited limited generalization to variable and evolving language patterns.

The advent of deep learning and Transformer architectures~\cite{devlin2019} brought substantial improvements. Domain-specialized variants such as \textit{BERTweet} for social media language~\cite{Nguyen2020BERTweet} and \textit{ClinicalBERT} for clinical text~\cite{Alsentzer2019ClinicalBERT} achieve 80--90\% accuracy in depression identification when trained on in-domain data~\cite{kelley2022,gadzama2024}. More recently, LLMs such as GPT-3.5 and GPT-4 have been explored in zero-shot and few-shot settings, demonstrating notable screening capability without any fine-tuning~\cite{ohse2024zero,Shin2024,Lho2025}. A fundamental limitation of LLMs, however, is the strong correlation between model size and performance, which entails significant computational costs for the largest models~\cite{ohse2024zero}.

\subsection{Emotions Associated with Depression}
\label{sec:related_emotions}

The literature has identified a recurring set of emotion categories in the language of individuals experiencing depression. The \emph{DepressionEmo} dataset~\cite{rahman2024depressionemo} provides an operational taxonomy of eight emotions annotated on Reddit posts: \textit{anger}, \textit{cognitive dysfunction}, \textit{emptiness}, \textit{hopelessness}, \textit{loneliness}, \textit{sadness}, \textit{suicide intent}, and \textit{worthlessness}. Recent studies confirm that these emotions do not operate as independent signals but tend to co-occur, forming a latent construct attributable to depressive distress~\cite{rahman2024depressionemo,tackman2019}.

\section{Methodology}
\label{sec:method}

\subsection{Depressive Severity Index}
\label{sec:index}

We propose a composite depressive severity index that exploits the emotion labels assigned by an LLM to the eight affective dimensions described in Section~\ref{sec:related_emotions}. Each emotion is treated as a binary variable (present~=~1, absent~=~0) and multiplied by a weight $w_i$ proportional to its clinical relevance. The resulting index is:

\begin{align}
S &= 1 \cdot \text{anger}
+ 1 \cdot \text{cog\_dysfunction} \notag \\
&\quad + 1 \cdot \text{emptiness}
+ 2 \cdot \text{hopelessness} \notag \\
&\quad + 1 \cdot \text{loneliness}
+ 1 \cdot \text{sadness} \notag \\
&\quad + 3 \cdot \text{suicide\_intent}
+ 2 \cdot \text{worthlessness}
\label{eq:index}
\end{align}

The weight assignments are grounded in clinical considerations: sadness, loneliness, anger, emptiness, and cognitive dysfunction are indicators of distress but non-specific in isolation ($w = 1$); hopelessness and worthlessness signal advanced suffering and are associated with worse prognosis ($w = 2$); suicide intent is the highest-criticality indicator and demands immediate attention ($w = 3$)~\cite{ribeiro2018depression,apa2022}.

The value $S$ is mapped onto four severity levels, modeled after the clinical thresholds of PHQ-9 and BDI-II:
\begin{itemize}
  \item $S = 0$--$1$: minimal or absent depression;
  \item $S = 2$--$4$: mild depression;
  \item $S = 5$--$6$: moderate depression;
  \item $S \geq 7$: severe depression (high alert).
\end{itemize}
The theoretical maximum score is $S = 13$ (all emotions present).

\subsection{Prompt Engineering}
\label{sec:prompt}

Emotional analysis is performed via \textit{prompt engineering}: each post is paired with a structured textual instruction that guides the LLM toward a consistent and machine-readable classification output. The base prompt for multi-label classification is:

\begin{lstlisting}[style=prompt]
Analyze the sentiment of the following
comment from Reddit: "{post}".
Classify which of the following emotions
apply: {emotions}.
Answer with a JSON object, with True or
False for each emotion.
\end{lstlisting}

where \texttt{\{post\}} is the comment text and \texttt{\{emotions\}} is the list of eight emotion labels. To incorporate direct score computation, the prompt is extended as follows:

\begin{lstlisting}[style=prompt]
Analyze the sentiment of the following
comment from Reddit: "{post}".
Classify which of the following emotions
apply: {emotions}.
Answer with a JSON object, with True or
False for each emotion. Then, compute a
severity score as follows: assign weights
to emotions (suicide_intent=3,
hopelessness=2, worthlessness=2,
cognitive_dysfunction=1, sadness=1,
emptiness=1, loneliness=1, anger=1).
Return also the field "severity_score"
with the sum of the weights for the
emotions classified as True.
\end{lstlisting}

An example of the model output is:
\begin{lstlisting}[style=prompt]
{
  "anger": false,
  "cognitive_dysfunction": true,
  "emptiness": false,
  "hopelessness": true,
  "loneliness": true,
  "sadness": true,
  "suicide_intent": false,
  "worthlessness": true,
  "severity_score": 7
}
\end{lstlisting}

All models are used in \textit{zero-shot} mode: no additional examples are provided beyond the prompt instructions. This choice allows us to assess the intrinsic capabilities of LLMs without any domain-specific training~\cite{Jin2025,ohse2024zero}.

\section{Experiments}
\label{sec:experiments}

We evaluated our approach across two distinct phases: a controlled benchmark and a large-scale, in-the-wild analysis. This section outlines the complete experimental framework used for these evaluations. First, we detail the datasets and experimental setup, which include the annotated \emph{DepressionEmo} corpus and a custom collection of nearly 470,000 recent mental-health-related Reddit posts. Next, we present the evaluated models, introducing the nine locally hosted large language models and the fine-tuned baselines selected for comparison. Finally, we define the evaluation metrics, specifically the micro- and macro-averaged variants of precision, recall, and F1-score used to comprehensively quantify multi-label classification performance.

\subsection{Datasets and Experimental Setup}

\paragraph{Controlled benchmark: DepressionEmo.}
The \emph{DepressionEmo} dataset~\cite{rahman2024depressionemo} contains $\approx6{,}000$ Reddit posts manually annotated with the eight emotions described in Section~\ref{sec:related_emotions} under a multi-label scheme. We adopt the original 80/20 train/validation split proposed by the authors and evaluate all LLMs exclusively on the validation partition, without any fine-tuning.

\paragraph{In-the-wild benchmark: Reddit 2024--2025.}
Using a custom-built web scraper, we collected posts from four mental-health-themed subreddits: \textit{r/anxiety}, \textit{r/depression}, \textit{r/depression\_partners}, and \textit{r/mentalhealth}, the same communities used to construct the DepressionEmo dataset~\cite{rahman2024depressionemo}. The collection spans January 2024 to May 2025, yielding a total of $\mathbf{469{,}692}$ posts ($318{,}000$ in 2024 and $151{,}692$ in the first half of 2025). The per-subreddit distribution is reported in Table~\ref{tab:dataset}.

\begin{figure}[t]
  \centering
  \includegraphics[width=1\linewidth]{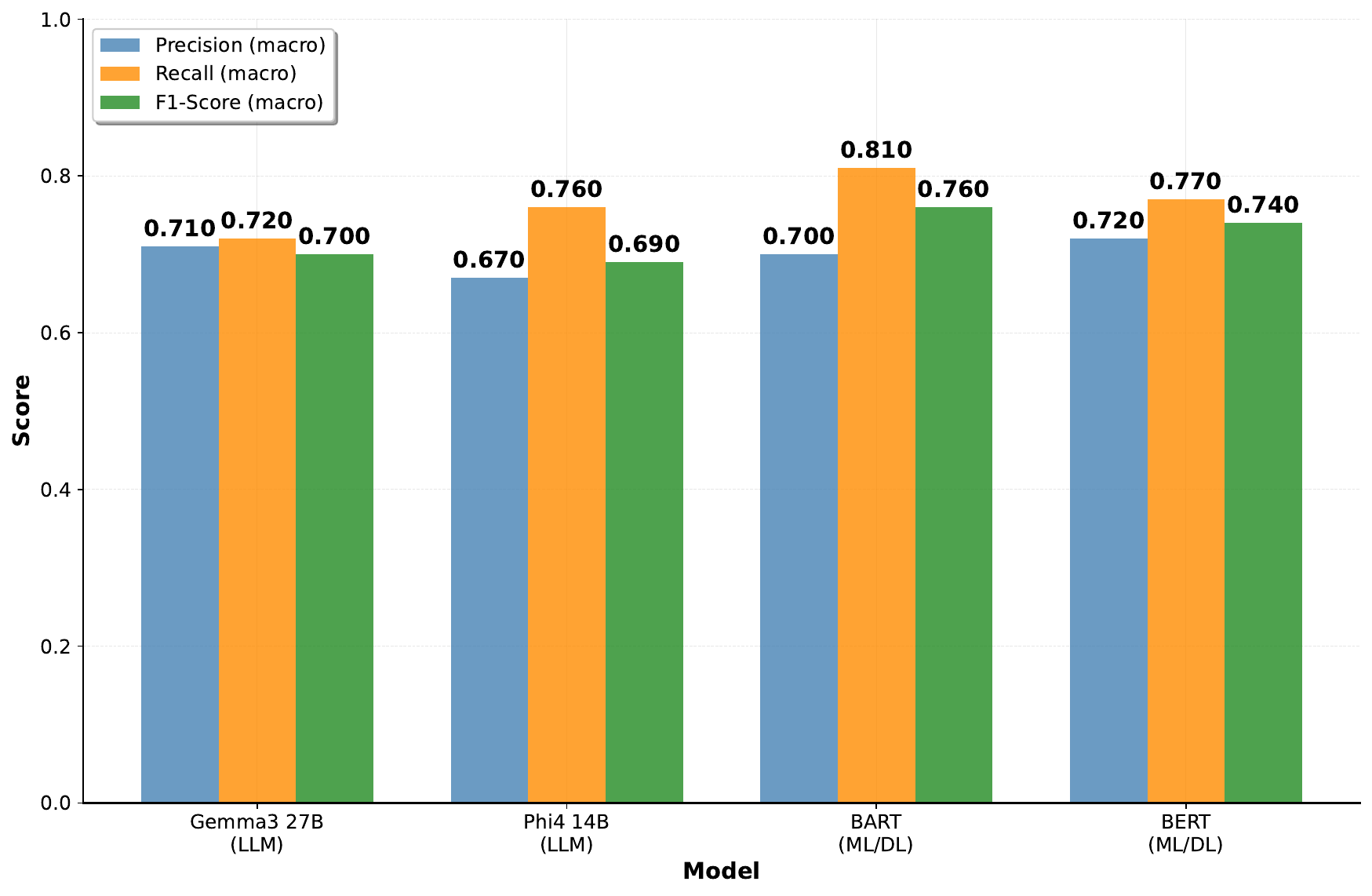}
  \caption{Comparison between LLMs (zero-shot) and fine-tuned models in terms of precision, recall, and F1-score. The two best models in each category are highlighted. LLMs trained on general-purpose data achieve competitive results relative to models specifically fine-tuned on depressive-domain data.}
  \label{fig:performance_comparison}
\end{figure}

\begin{table}[b]
\centering
\caption{Post distribution per subreddit in the in-the-wild dataset.}
\label{tab:dataset}
\small
\begin{tabular}{lrrr}
\toprule
\textbf{Subreddit} & \textbf{2024} & \textbf{2025 (H1)} & \textbf{Total} \\
\midrule
r/anxiety           & 96{,}068  & 42{,}660  & 138{,}728 \\
r/depression        & 50{,}276  & 113{,}314 & 163{,}590 \\
r/depr.\_partners   & 1{,}569   & 878       & 2{,}447   \\
r/mentalhealth      & 107{,}049 & 57{,}878  & 164{,}927 \\
\midrule
\textbf{Total}      & \textbf{318{,}000} & \textbf{151{,}692} & \textbf{469{,}692} \\
\bottomrule
\end{tabular}
\end{table}

\subsection{Evaluated Models}

In the controlled benchmark phase, we compare nine LLMs of varying sizes, run \textbf{locally} via the \textit{Ollama} framework, against literature baselines (SVM, LightGBM, XGBoost, GAN-BERT, BERT, BART) fine-tuned on the 80\% training split of DepressionEmo. The evaluated LLMs are: \texttt{qwen3:0.6b}, \texttt{phi3:mini}, \texttt{gemma2:2b}, \texttt{llama3.2:3b}, \texttt{phi4:14b}, \texttt{mistral:7b}, \texttt{samantha-mistral:7b}, \texttt{qwen3:14b}, and \texttt{gemma3:27b}. For the in-the-wild analysis, we employ the best-performing model identified in the benchmark, namely \texttt{gemma3:27b}.

\subsection{Evaluation Metrics}

Multi-label classification performance is evaluated using \textit{precision}, \textit{recall}, and \textit{F1-score}, computed in both the \textbf{micro} variant (globally aggregated counts across all classes) and the \textbf{macro} variant (arithmetic mean of per-class metrics). Their joint use enables assessment of both overall performance and behavior on less frequent classes, a critical consideration in the mental health domain~\cite{rahman2024depressionemo}.

\section{Results and Discussion}
\label{sec:results}

\subsection{Controlled Benchmark}

Full results of the controlled benchmark are reported in Table~\ref{tab:results} and visualized in Figure~\ref{fig:performance_comparison}.

\begin{table*}[ht]
\centering
\caption{Comparison of zero-shot LLMs and fine-tuned literature models on the DepressionEmo dataset (micro and macro precision, recall, F1-score). Best results per metric in bold. \cmark = fine-tuned; \xmark = zero-shot.}
\label{tab:results}
\begin{tabular}{l c ccc ccc}
\toprule
\textbf{Method} & \textbf{FT} &
\multicolumn{3}{c}{\textbf{Micro}} &
\multicolumn{3}{c}{\textbf{Macro}} \\
\cmidrule(lr){3-5}\cmidrule(lr){6-8}
 & & Prec. & Rec. & F1 & Prec. & Rec. & F1 \\
\midrule
SVM~\cite{rahman2024depressionemo}       & \cmark & 0.77 & 0.51 & 0.61 & 0.72 & 0.41 & 0.47 \\
LightGBM~\cite{rahman2024depressionemo}  & \cmark & 0.52 & 0.86 & 0.65 & 0.48 & 0.80 & 0.58 \\
XGBoost~\cite{rahman2024depressionemo}   & \cmark & 0.69 & 0.63 & 0.66 & 0.63 & 0.56 & 0.59 \\
GAN-BERT~\cite{rahman2024depressionemo}  & \cmark & 0.73 & 0.77 & 0.75 & 0.69 & 0.72 & 0.70 \\
BERT~\cite{rahman2024depressionemo}      & \cmark & 0.76 & 0.83 & 0.79 & 0.72 & 0.77 & 0.74 \\
BART~\cite{rahman2024depressionemo}      & \cmark & 0.74 & 0.86 & \textbf{0.80} & 0.70 & 0.81 & \textbf{0.76} \\
\midrule
qwen3:0.6b   & \xmark & \textbf{0.83} & 0.33 & 0.47 & \textbf{0.86} & 0.26 & 0.35 \\
phi3:mini    & \xmark & 0.63 & 0.85 & 0.72 & 0.64 & 0.78 & 0.67 \\
gemma2:2b    & \xmark & 0.58 & 0.87 & 0.70 & 0.61 & 0.84 & 0.65 \\
llama3.2:3b  & \xmark & 0.65 & 0.83 & 0.73 & 0.63 & 0.77 & 0.67 \\
phi4:14b     & \xmark & 0.68 & 0.82 & 0.74 & 0.67 & 0.76 & 0.69 \\
mistral:7b   & \xmark & 0.66 & 0.84 & 0.74 & 0.65 & 0.76 & 0.68 \\
samantha-mistral:7b & \xmark & 0.56 & \textbf{0.90} & 0.69 & 0.56 & \textbf{0.86} & 0.65 \\
qwen3:14b    & \xmark & 0.64 & 0.84 & 0.73 & 0.65 & 0.77 & 0.67 \\
gemma3:27b   & \xmark & 0.73 & 0.77 & 0.75 & 0.71 & 0.72 & 0.70 \\
\bottomrule
\end{tabular}
\end{table*}

Results show that the gap between zero-shot LLMs and fine-tuned models is modest: \texttt{gemma3:27b} achieves micro-F1 = 0.75 and macro-F1 = 0.70, matching fine-tuned GAN-BERT and falling only 0.05 F1 points below BERT and BART. Overall it's the more balanced model among LLMs. The performance decrease relative to fine-tuned models is expected, yet remains small given the complete absence of domain-specific training.

\paragraph{Per-model analysis.}
\texttt{gemma3:27b} is the best-performing LLM overall, exhibiting the lowest divergence between precision (0.73) and recall (0.77), indicative of balanced behavior. \texttt{qwen3:0.6b} achieves very high precision (0.83) but very low recall (0.33), reflecting a conservative labeling strategy. Conversely, \texttt{samantha-mistral:7b}, a psychologically-adapted variant of Mistral, maximizes recall (0.90) at the expense of precision (0.56), a trade-off consistent with prevention scenarios where the cost of false negatives is high, but with a precision particularly low. In fact, the F1 score it's lower than other more powerful models (\textit{i.e.}, gemma3:27b).

\paragraph{Clinical implications.}
In screening and prevention contexts, minimizing false negatives (i.e., undetected critical cases) is the primary objective, making high recall a desirable property. Larger models generally show better precision--recall balance, suggesting that model scale is the primary predictor of performance, consistent with~\cite{ohse2024zero}.

\paragraph{Advantage of LLMs over fine-tuned models.}
A key advantage is the ability of LLMs to operate without labeled data: fine-tuned models require domain-specific annotations and risk becoming obsolete as online language evolves. LLMs, having been exposed to vast quantities of general text, generalize more naturally to novel expressions of depressive distress~\cite{Jin2025}.

\subsection{In-the-Wild Analysis}

\subsubsection{Emotional Structure of Communities}

\begin{figure}[t]
  \centering
  \includegraphics[width=\linewidth]{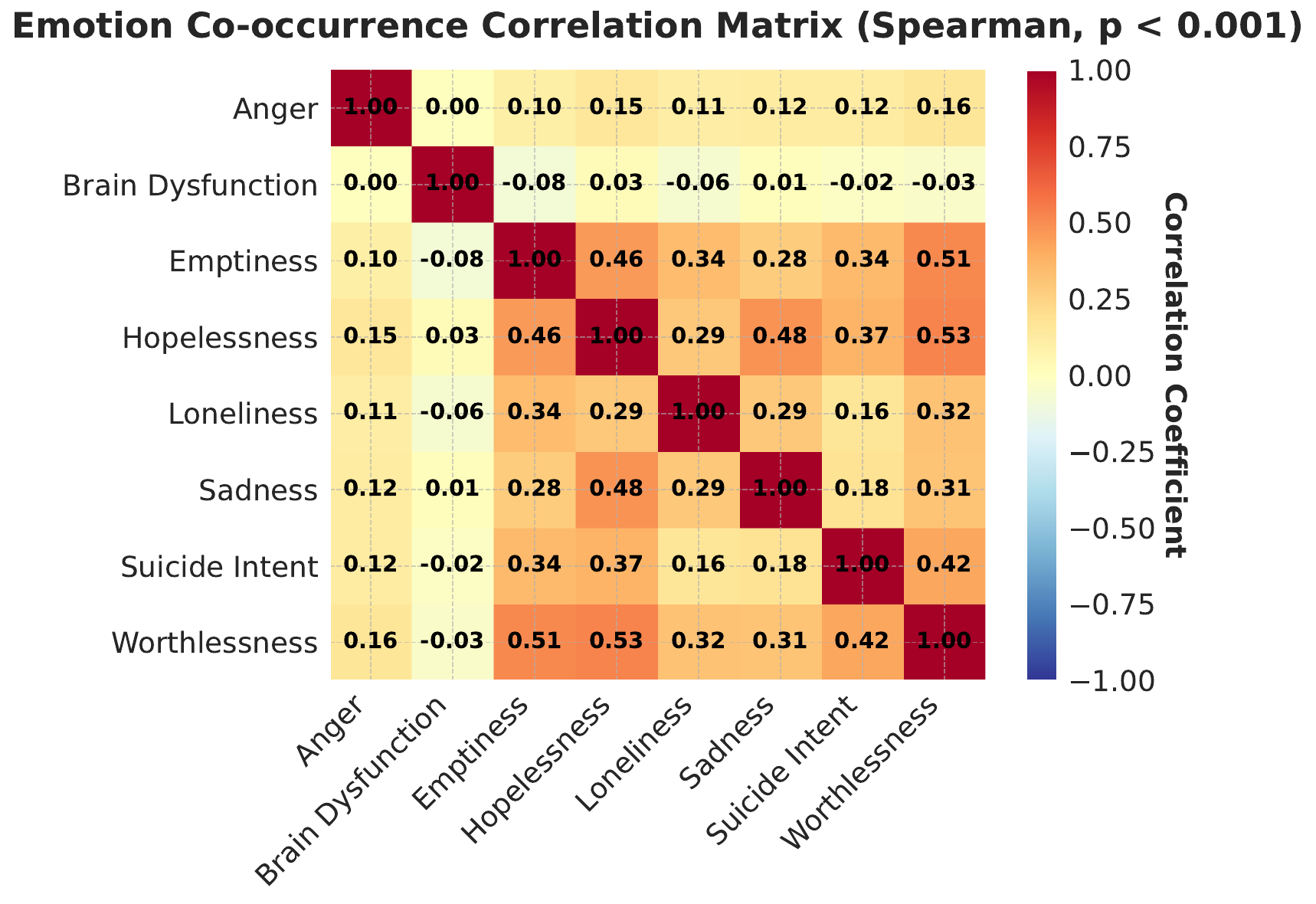}
  \caption{Spearman correlation matrix of detected emotions across the full in-the-wild dataset ($p < 0.001$). Emotions belonging to the depressive core (\textit{hopelessness}, \textit{worthlessness}, \textit{sadness}, \textit{emptiness}) exhibit significant positive correlations ($\rho = 0.28$--$0.53$).}
  \label{fig:correlation}
\end{figure}

Figure~\ref{fig:correlation} displays the Spearman correlation matrix of detected emotions. \textit{Hopelessness}, \textit{worthlessness}, \textit{sadness}, and \textit{emptiness} show moderate to high positive correlations ($\rho \approx 0.28$--$0.53$, $p < 0.001$), indicating high co-occurrence within the same posts. This pattern supports the hypothesis that these emotions are observable manifestations of a shared latent construct corresponding to depressive distress.

Conversely, \textit{anger} and \textit{cognitive dysfunction} show near-zero correlations with most other categories, suggesting weaker internal coherence with the depressive core and higher discriminant validity relative to the primary domain.

\begin{figure}[t]
  \centering
  \includegraphics[width=\linewidth]{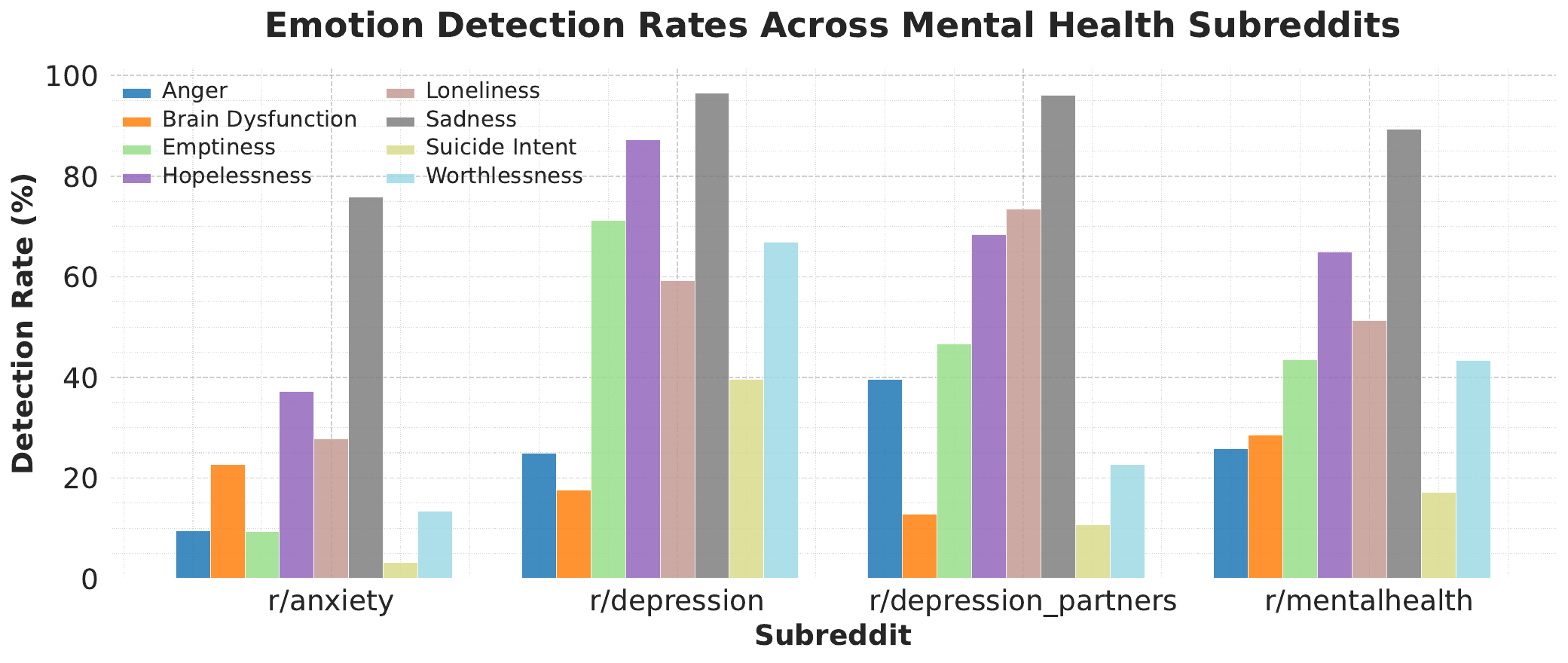}
  \caption{Emotion detection rates across the four subreddits. In \textit{r/depression}, \textit{sadness} and \textit{hopelessness} exceed 80--90\% of posts; in \textit{r/anxiety}, the same emotions appear at significantly lower frequencies.}
  \label{fig:emotion_dist}
\end{figure}

Figure~\ref{fig:emotion_dist} shows how emotional profiles vary substantially across communities. In \textit{r/depression}, \textit{sadness} and \textit{hopelessness} are detected in the vast majority of posts (often $>80$--$90$\%), whereas in \textit{r/anxiety} the same emotions are considerably less frequent, outlining a distinct affective profile consistent with the differential nature of the two disorders.

\subsubsection{Risk Score Distributions}

\begin{figure}[t]
  \centering
  \includegraphics[width=\linewidth]{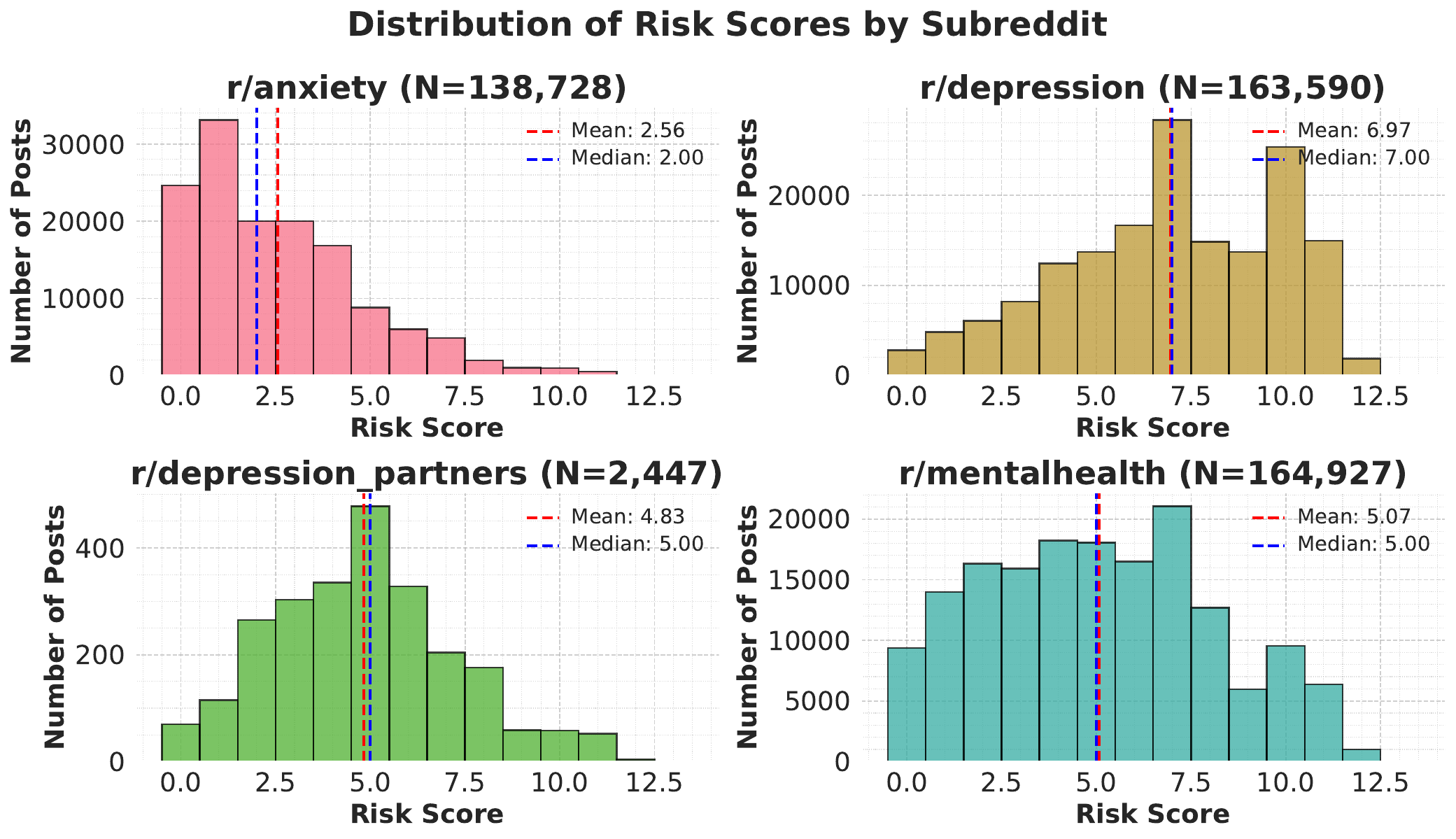}
  \caption{Distributions of risk scores $S$ (Eq.~\ref{eq:index}) per subreddit. In \textit{r/depression} the distribution is shifted toward higher values, with mean and median nearly coinciding ($S \approx 7$); in \textit{r/anxiety} the distribution is strongly concentrated at low values (median~$\approx$~2).}
  \label{fig:risk_dist}
\end{figure}

\begin{figure}[t]
  \centering
  \includegraphics[width=\linewidth]{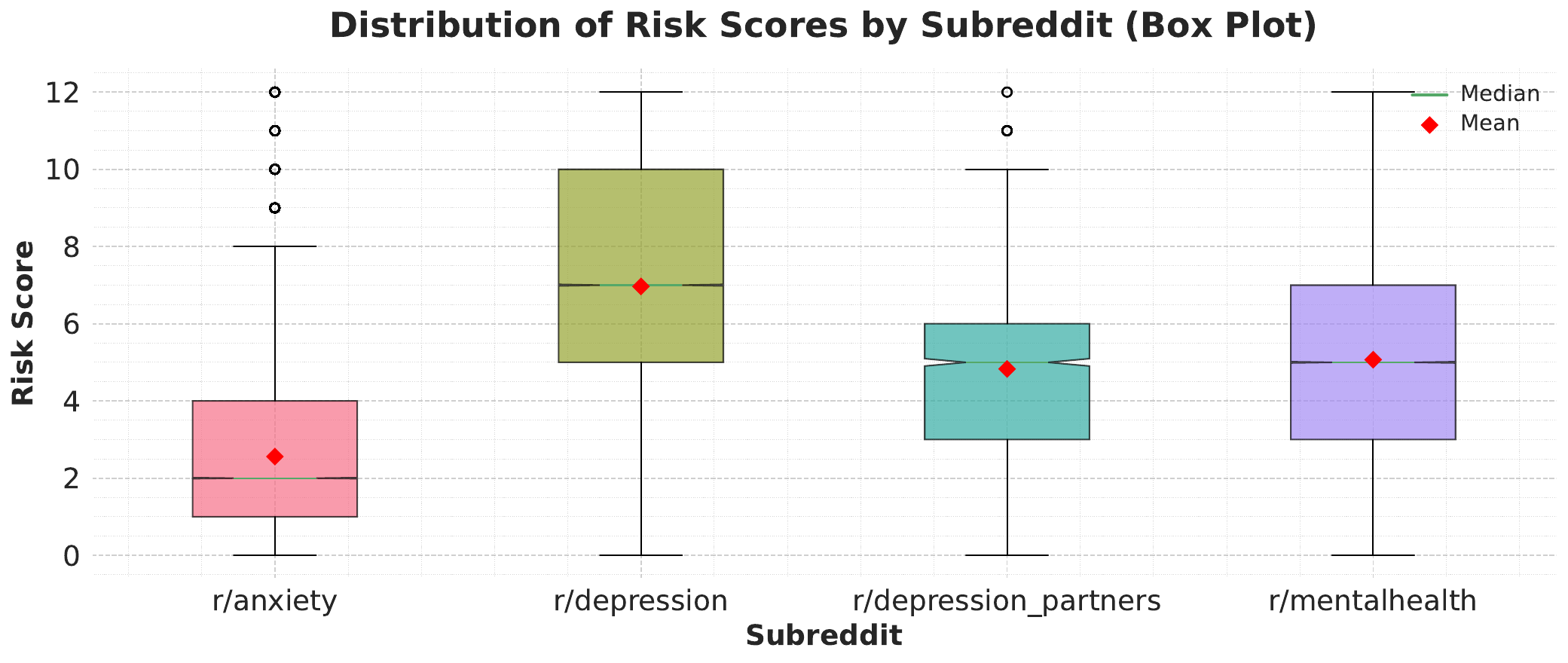}
  \caption{Box plots of risk scores per subreddit. Significant differences emerge across communities; \textit{r/depression} exhibits both a higher median and greater variability at the extreme values.}
  \label{fig:boxplot}
\end{figure}

Figures~\ref{fig:risk_dist} and~\ref{fig:boxplot} demonstrate that $S$ effectively discriminates between communities. In \textit{r/depression}, the distribution is shifted toward high values (mean $\approx 7$, approximately 43\% of posts $\geq 7$), suggesting a structurally elevated and homogeneously distributed risk. In \textit{r/anxiety}, the distribution is strongly left-skewed (median $\approx 2$, only $\approx 3$\% of posts $\geq 7$), consistent with anxiety rather than depressive profiles. The subreddits \textit{r/depression\_partners} and \textit{r/mentalhealth} exhibit intermediate profiles with greater variability.

\subsubsection{High-Risk Post Analysis}

\begin{figure}[t]
  \centering
  \includegraphics[width=\linewidth]{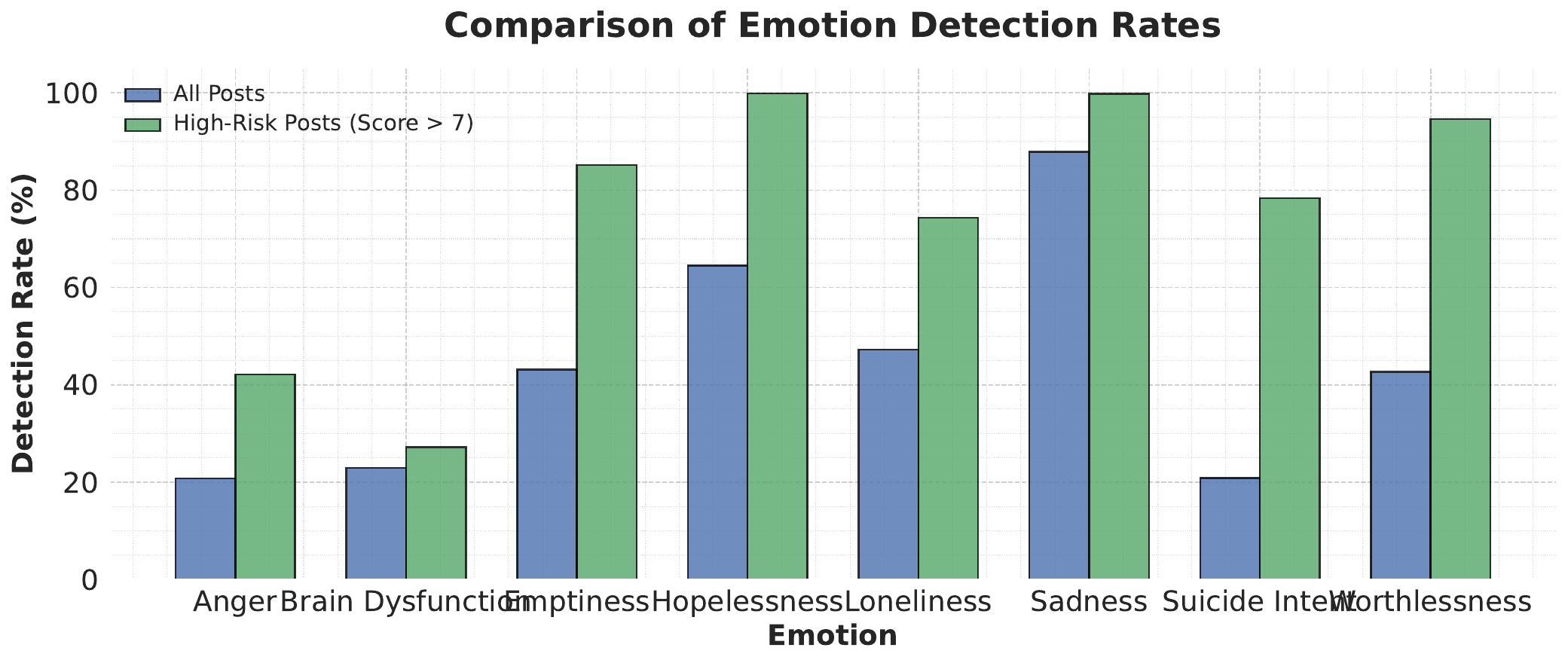}
  \caption{Comparison of emotion detection rates between the full post set and the high-risk subset ($S > 7$). In high-risk posts, \textit{sadness} and \textit{hopelessness} are detected in nearly all cases; a marked increase in \textit{worthlessness}, \textit{emptiness}, and \textit{suicide intent} is also observed.}
  \label{fig:high_risk}
\end{figure}

Figure~\ref{fig:high_risk} confirms that high values of $S$ are not attributable to random fluctuations, but reflect the systematic convergence of multiple critical affective indicators. In posts with $S > 7$, \textit{suicide intent}, which carries the maximum weight of 3, is markedly overrepresented relative to the global distribution, validating the index's ability to isolate the most critical cases.

\subsubsection{Temporal Analysis}

\begin{figure}[t]
  \centering
  \includegraphics[width=\linewidth]{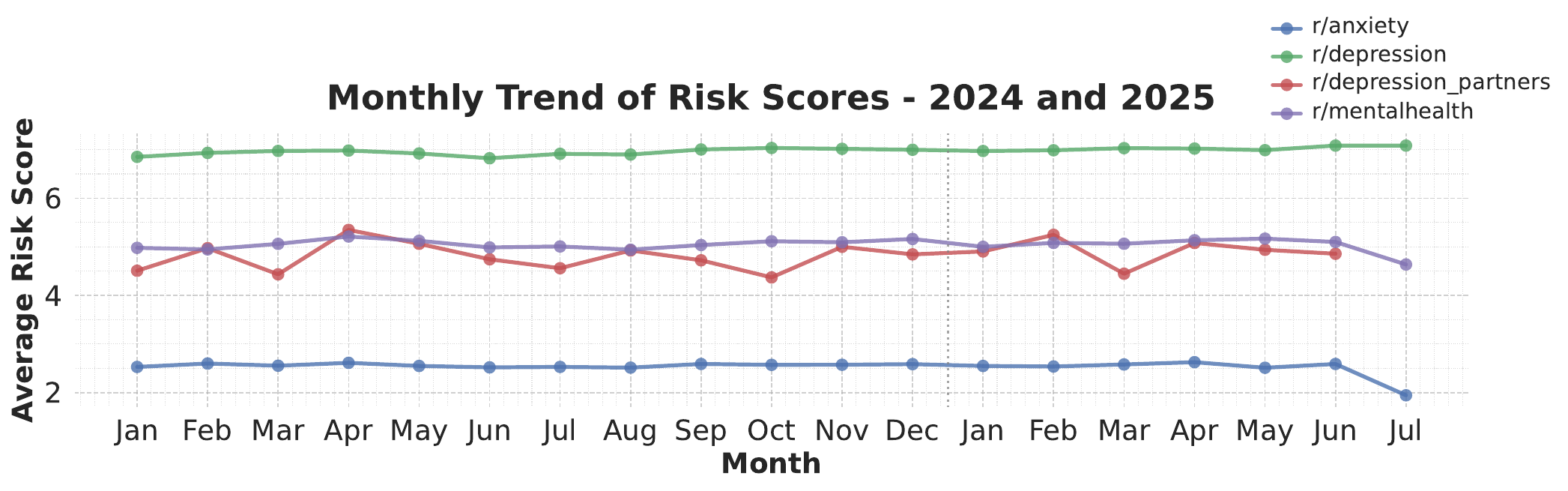}
  \caption{Monthly average risk score from January 2024 to July 2025. Values remain stable over time per subreddit, with structural inter-community differences persisting throughout the observation period. A moderate upward trend is observable in the first half of 2025.}
  \label{fig:monthly_trends}
\end{figure}

\begin{figure}[t]
  \centering
  \includegraphics[width=\linewidth]{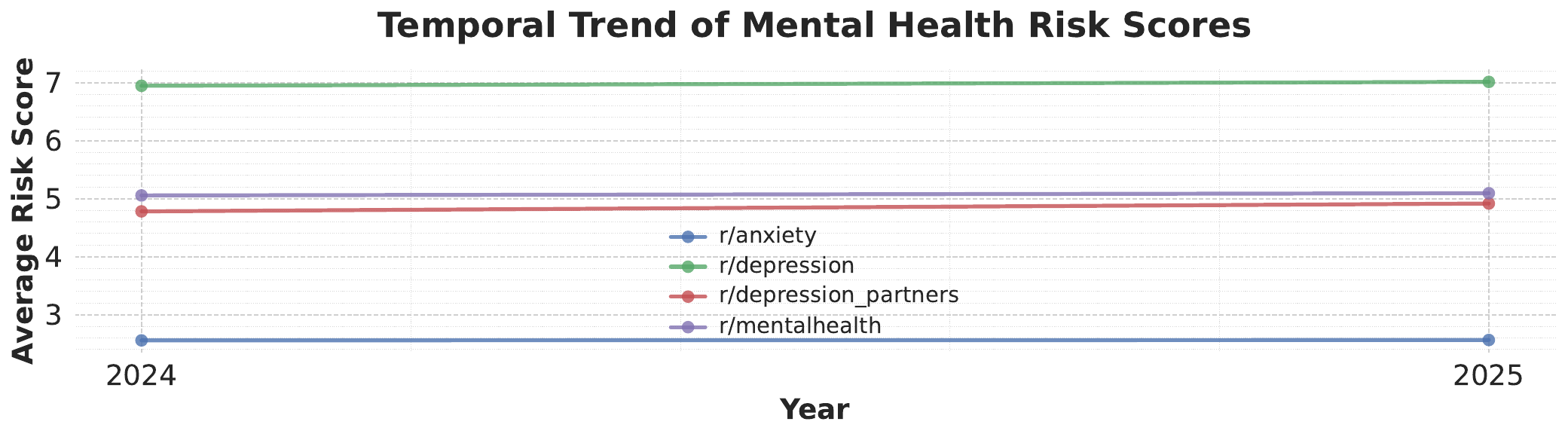}
  \caption{Fine-grained temporal evolution of the mean risk score. The absence of erratic oscillations indicates that the system does not overreact to episodic language variation.}
  \label{fig:temporal_trends}
\end{figure}

Figures~\ref{fig:monthly_trends} and~\ref{fig:temporal_trends} reveal substantial temporal stability of risk scores throughout 2024--2025, with no marked discontinuities. This stability is a relevant property for a monitoring system: it indicates that the model responds to structured linguistic variation rather than episodic noise.

A moderate increase in mean scores during the first half of 2025 is observable in several subreddits. While causal inference is not possible, this trend may reflect an intensification of depressive language associated with broader social or contextual factors. This reading is consistent with the value of a longitudinal monitoring system capable of detecting drifts in collective language that may carry psychologically meaningful signals~\cite{Chancellor2020,Naslund2019}.

\subsubsection{Limitations}

This work has several methodological limitations. First, the analysis relies exclusively on public text data and does not account for individual or clinical variables of the post authors, limiting the generalizability of estimates at the individual level. Second, emotion interpretation depends on the semantic capabilities of the LLM, which, however advanced, remain a simplification of the underlying psychological process. Finally, the self-referential nature of Reddit content introduces potential self-selection bias: the monitored communities tend to attract users with already-manifest distress, limiting the representativeness of the sample with respect to the general population.

From a system perspective, the gap of approximately 0.05--0.06 F1 points relative to fine-tuned models leaves room for improvement via hybrid approaches (few labeled examples for few-shot prompting) or lightweight fine-tuning (LoRA/QLoRA) on smaller models. Ethical considerations regarding privacy and informed consent require careful attention before deploying such tools in real-world applicative contexts.

\section{Conclusions}
\label{sec:conclusions}

We have presented an LLM-based system for automated depression risk assessment in social media, centered on a clinically-grounded weighted severity index derived from eight emotionally relevant dimensions. Evaluation on the \emph{DepressionEmo} benchmark demonstrates that large locally-run models, in particular \texttt{gemma3:27b}, achieve performance competitive with purpose-built fine-tuned models (micro-F1 = 0.75 vs. 0.80 for BART), operating in zero-shot mode without any domain-specific training.

The in-the-wild analysis on $469{,}692$ Reddit posts confirms the ecological validity of the approach: the system produces coherent and interpretable risk profiles, successfully differentiating semantically distinct communities (e.g., \textit{r/depression} vs. \textit{r/anxiety}) and maintaining stable estimates over time. The proposed index enables automatic identification of high-risk posts through the systematic convergence of multiple critical affective indicators, providing a scalable psychological triage tool.

The main contributions of this work are: (1) the definition of a composite depressive severity index inspired by standardized clinical scales but adapted for automated text analysis; (2) the demonstration that medium-to-large LLMs, runnable locally without specialized hardware, can achieve meaningful performance in depressive emotion recognition; (3) the first large-scale longitudinal analysis of depression risk in major mental-health subreddits over the 2024--2025 period.

Our findings suggest that LLMs, by virtue of their domain-agnostic training and ability to generalize to evolving linguistic expressions, represent a promising resource for large-scale psychological monitoring, with infrastructure costs substantially lower than proprietary API-based models. The proposed system is designed to complement rather than replace clinical judgment, serving as a first-pass filter in continuous monitoring scenarios. Future work will include extension to non-English languages, integration of behavioral signals (post frequency and timing), and validation against clinical samples to calibrate index thresholds against certified diagnoses.

\printbibliography[heading=bibintoc, title={References}]

\end{document}